\documentclass[11pt]{article}

\usepackage[T1]{fontenc}
\usepackage[utf8]{inputenc}
\usepackage{lmodern}
\usepackage{textcomp}
\usepackage{amsmath,amsfonts,amssymb}
\usepackage{booktabs}
\usepackage{tabularx}
\usepackage{multirow}
\usepackage{algorithm}
\usepackage{algorithmic}
\usepackage{array}
\usepackage{stfloats}
\usepackage{url}
\usepackage{verbatim}
\usepackage[percent]{overpic}
\usepackage{graphicx}
\usepackage{subfig}
\usepackage{float}
\usepackage{cite}
\usepackage[a4paper,margin=1in]{geometry}
\usepackage{authblk}
\usepackage{microtype}
\usepackage[hidelinks]{hyperref}

\graphicspath{{figures/}}

\newcommand{\mthreeedseq}{\texttt{falcon\_\allowbreak indoor\_\allowbreak flight\_\allowbreak 1}}

\makeatletter
\let\arxiv@orig@includegraphics\includegraphics
\renewcommand{\includegraphics}[2][]{%
  \IfFileExists{#2}{\arxiv@orig@includegraphics[#1]{#2}}{%
    \IfFileExists{figures/#2}{\arxiv@orig@includegraphics[#1]{figures/#2}}{%
      \begingroup
      \setlength{\fboxsep}{6pt}%
      \fbox{\begin{minipage}[c][0.18\linewidth][c]{0.78\linewidth}\centering\small Missing figure file: \texttt{\detokenize{#2}}\end{minipage}}%
      \endgroup
    }%
  }%
}
\makeatother

\title{Minimal Solvers for Full-DoF Motion Estimation from Asynchronous Differential SfM}

\author[1,2]{Shuo Pan}
\author[1,2]{Banglei Guan\thanks{Corresponding author: \texttt{guanbanglei12@nudt.edu.cn}}}
\author[1,2]{Bin Li}
\author[1,2]{Zhenbao Yu}
\author[1,2]{Zibin Liu}
\author[1,2]{Zi Wang}
\author[1,2]{Yang Shang}
\author[1,2]{Qifeng Yu}

\affil[1]{College of Aerospace Science and Engineering, National University of Defense Technology, Changsha 410073, China}
\affil[2]{The Hunan Provincial Key Laboratory of Image Measurement and Vision Navigation, Changsha 410073, China}

\date{}

\begin{document}

\maketitle

\begin{abstract}
As a bio-inspired intelligent sensor, event cameras have introduced a new paradigm in the intelligent perception of spatiotemporal information and visual motion estimation, characterized by their high temporal resolution, low latency, and minimal power consumption. However, their asynchronous data streams present significant challenges to traditional synchronous, frame-based algorithms. To address these challenges, this paper presents a novel framework for full degree of freedom (DoF) egomotion estimation directly from asynchronous optical flow, specifically targeting the joint recovery of angular and linear velocities. We decouple the differential epipolar constraint into distinct angular and linear velocity components, and derive its formulation for asynchronous data. Based on this formulation, an optimization algorithm is developed that enables full-DoF egomotion estimation leveraging at least five points. Furthermore, by applying a first-order approximation to rotational dynamics, we transform the constraint equations into a polynomial form, resulting in the first algebraic minimal 5-point solver for this formulation. To ensure real-time performance in high-speed scenarios, we additionally propose an accelerated solver achieved by truncating high-order angular velocity terms. Extensive evaluations on both synthetic and real-world datasets demonstrate that the asynchronous approach outperforms traditional synchronous methods, particularly in its accuracy and robustness to spatiotemporal noise. We believe that this work establishes a critical foundation for efficient and accurate continuous-time motion estimation in high-speed robotics applications.
\end{abstract}


\section{Introduction}

Accurate egomotion estimation is fundamental to equipping unmanned systems with autonomous navigation capabilities in critical applications, including autonomous driving, unmanned aerial vehicle control, and visual servoing~\cite{selfdrive,UAV,v_s,embody}. Inspired by biological vision mechanisms, event cameras, with their unique combination of high temporal resolution, exceptional dynamic range, and low power consumption, have emerged as a promising technology, attracting broad interest and demonstrating considerable potential across numerous fields~\cite{eventsapply1,eventsapply2,eventsapply3}. Consequently, achieving fast and accurate egomotion estimation with event cameras has recently become a key research focus~\cite{survey1,survey2}.

Unlike the conventional imaging paradigm~\cite{frame1,frame2}, event cameras asynchronously report per-pixel brightness changes as a stream of events~\cite{cmos}. This asynchronicity calls for a paradigm shift in algorithm design, from synchronous processing to methods that natively exploit the spatiotemporal structure of events for rapid estimation of motion~\cite{asynchronous1,asynchronous2}. Previous works on monocular egomotion estimation with event cameras have typically followed conventional imaging approaches, accumulating events into frames~\cite{Cmax1,Dmin,STalig} and applying conventional synchronous-vision algorithms, such as those based on feature matching, geometric constraints, optical flow, deep learning, or using sensors requiring extra data~\cite{AC1, AC2, 5pts,geo,opticalflow,egolearning,line1,line2}. These strategies inevitably constrain the intrinsic high-speed capabilities and independence of event cameras. To overcome these limitations, some methods have adopted event-native strategies directly to estimate motion parameters from the event stream. These include leveraging spatial features like lines~\cite{line3}, normal flow~\cite{normflow}, as well as increasingly sophisticated data-driven techniques that have demonstrated considerable promise in recent years~\cite{snnego}. A notable attempt to unify the modeling of rolling shutter cameras and event cameras under a common asynchronous framework is presented in~\cite{asynchronous1}, which presented the minimal solvers for egomotion estimation via feature matching under the asynchronous framework.

However, the performance of feature matching in event-based vision is often limited~\cite{survey1}. Rather than capturing absolute photometric intensity, event cameras asynchronously trigger sparse events primarily along brightness-changing edges~\cite{ev}. This intrinsic sensing modality inherently destabilizes the feature extraction and matching processes upon which motion estimation relies, thereby presenting a fundamental bottleneck for event-based egomotion algorithms. To break through this fundamental bottleneck, computing optical flow from event streams has become a crucial research trajectory. By providing a low-latency, dense estimation of velocity fields directly from brightness changes, optical flow naturally facilitates more resilient egomotion tracking without the need for fragile point matching. This paradigm is fundamentally aligned with the asynchronous streaming nature of event data, as it can natively interpret the spatiotemporal correlations within events and fully leverage their microsecond-level timestamp precision. These inherent advantages have rendered optical flow estimation a fundamental primitive in event-based vision, serving as a robust precursor for precise egomotion estimation~\cite{EMoflow,EV-FlowNet,ERAFT,STE,ADM,gra,evf,snn1,snn2,hug,dense}.
\begin{figure}[htbp]
	\centering
	\includegraphics[width=0.7\linewidth]{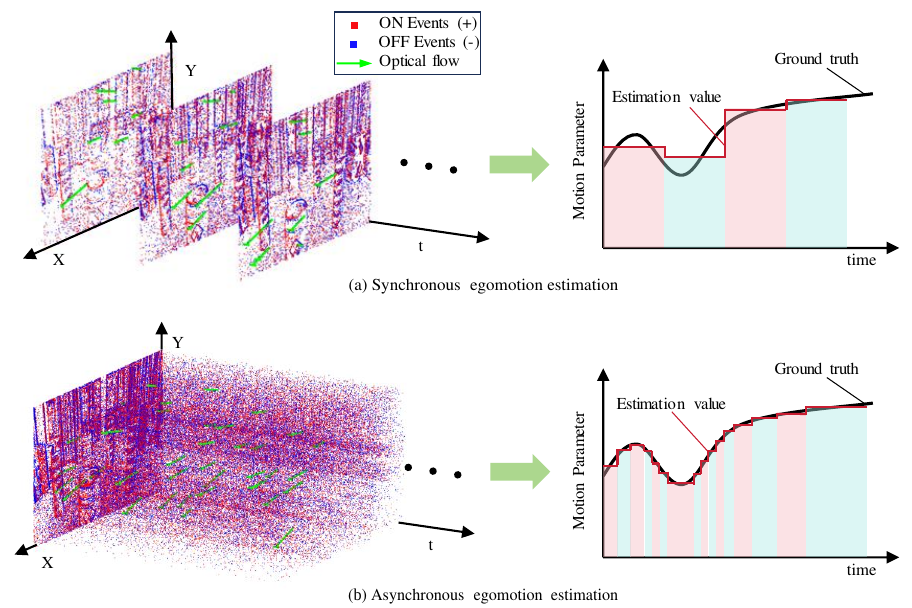}
	\caption{Comparison between traditional synchronous and asynchronous motion estimation. (a) Frame-based synchronous estimation methods inherently discretize temporal resolution with fixed steps, resulting in latency and piecewise-constant estimation errors. (b) Estimating egomotion parameters directly from asynchronous optical flow enables motion estimation over adaptive time intervals, allowing the estimates to closely track the true motion dynamics.}
	\label{fig:taster}
\end{figure}

As illustrated in Figure \ref{fig:taster}, when performing ego-motion estimation via differential epipolar constraints, traditional synchronous methods operate on fixed time intervals, inherently suffering from temporal discretization artifacts and latency, leading to piecewise-constant approximations that fail to capture dynamic motion. In contrast, the asynchronous paradigm directly integrates event-driven optical flow over adaptive time horizons, enabling continuous-time estimation that precisely tracks the true motion dynamics with high temporal fidelity. Nevertheless, there is a lack of the systematic modeling and solving of asynchronous differential epipolar constraints. The objective of this paper is to propose a novel egomotion estimation method for event cameras, building upon the extraction of asynchronous optical flow. In this paper, we establish an asynchronous formulation of the differential epipolar geometry constraints, and derive iterative solver and minimal algebraic solvers for the egomotion estimation of a monocular event camera. To the best of our knowledge, this is the first minimal algebraic solver proposed for the differential epipolar constraints. The primary contributions of this paper can be summarized as follows:

$ \bullet \, $ \textbf{A Novel Asynchronous and Decoupled Formulation}: We establish a new framework for full-DoF monocular egomotion estimation from asynchronous optical flow, grounded in differential epipolar geometry. This novel constraint formulation naturally decouples rotational and translational motion parameters. By explicitly modeling geometric constraints across different timestamps, it provides a more tractable and stable foundation for subsequent optimization compared to traditional coupled approaches.

$ \bullet \, $ \textbf{Iterative Solver Based on Eigenvalue Minimization}: Building upon the decoupled model, we develop an iterative optimization solver. This solver estimates motion parameters by sequentially minimizing the smallest eigenvalue of a data matrix constructed from the constraints. It operates on a minimal set of only 5 event-flow correspondences and achieves full-DoF motion estimation.

$ \bullet \, $ \textbf{Minimal Gr\"obner Basis-based Solvers via Polynomialization}: By employing a first-order approximation for rotation, we transform the problem into a polynomial system and derive a closed-form solution via the Gr\"obner basis method, yielding the first-of-its-kind minimal solver. Furthermore, through strategically truncating higher-order angular velocity terms, we present an accelerated variant of this solver, which reduces computational complexity and achieves nearly an order of magnitude speedup with only a marginal loss in precision. 

This paper is organized as follows: 
\textbf{Sec.\ref{sec:Related works}} provides the necessary background on differential epipolar constraints and reviews related works in event-based optical flow and motion estimation.
\textbf{Sec.\ref{sec:Problem formulation}} formulates the problem, introducing the decoupled differential epipolar constraints and extending it to the asynchronous event stream.
\textbf{Sec.\ref{sec: Egomotion Estimation}} presents three minimal solvers.
\textbf{Sec.\ref{sec:Experiment}} describes the experimental setup, including the synthetic and real-world datasets.
\textbf{Sec.\ref{sec: Conclusion}} concludes this paper.

\section{Related works}
\label{sec:Related works}
\textbf{Differential epipolar constraints}: The differential epipolar constraints establish a fundamental relationship between camera motion parameters (linear and angular velocities) and optical flow, enabling the estimation of relative pose and finding broad applications in visual odometry~\cite{vod}, 3D reconstruction~\cite{d3dd}, and image correction~\cite{aware}. Early work by Ma et al.~\cite{invitation} laid the theoretical groundwork. Subsequent research has focused on developing efficient solvers for various camera models. For global-shutter cameras, a linear 8-point solver was proposed~\cite{8pts} to recover motion parameters. 
To correct image distortion and motion parameter errors in rolling-shutter cameras, solvers were developed under both constant velocity and constant acceleration assumptions~\cite{aware}. More recent advancements include introducing radial distortion into the Wconstraint framework, accompanied by a robust generalized eigenvalue solver and a maximum-likelihood refinement scheme~\cite{distortion}, and extending the formulation to generalized rolling-shutter stereo systems with corresponding solving strategies~\cite{rig}. The recently proposed E-MoFlow by Li et al.~\cite{EMoflow} establishes a novel, unsupervised framework for joint optical flow and egomotion estimation. By integrating differential epipolar constraints, the method bypasses explicit depth estimation and achieves superior, state-of-the-art performance on multiple benchmarks.

\textbf{Event-based optical flow estimation}: Diverging from frame-based techniques, the input of optical flow estimation in event-based vision is no longer a sequence of continuous frames but rather sparse, unstructured spatiotemporal point clouds. Consequently, classical optical flow algorithms, such as Lucas-Kanade~\cite{LK} and Horn-Schunck~\cite{HS}, cannot be directly applied. Due to the asynchronous nature of event cameras, processing methods can be categorized as synchronous if they force continuous event streams into fixed time slices, and asynchronous if they process events individually according to their independent arrival times.

Since modern GPUs and deep learning frameworks are optimized for dense matrix operations, synchronous processing of event streams has become a pragmatic yet efficient strategy. Representative works include EV-FlowNet~\cite{EV-FlowNet}, which estimates flow on discrete event voxels via self-supervised learning, and E-RAFT~\cite{ERAFT}, which adapts the RAFT architecture using correlation volumes and recurrent units to achieve high-precision predictions in complex scenes. To enhance spatiotemporal modeling, STE-FlowNet~\cite{STE} introduces a spatiotemporal recurrent encoder, while ADM-Flow~\cite{ADM} addresses information loss during discretization through an Adaptive Density Module. These methods have demonstrated superior performance on multiple datasets.

Asynchronous optical flow algorithms aim to exploit the high temporal resolution of event streams. Early gradient-based methods utilized local differentiation of the event surface or plane fitting to resolve motion direction~\cite{gra}~\cite{evf}. Within deep learning, spiking neural networks (SNNs) have become the dominant asynchronous architecture; for instance, Spike-FlowNet~\cite{snn1} combines SNNs with ANNs to handle sparse spike inputs, while hierarchical SNN models~\cite{snn2} utilize spike synchrony for estimation. Additionally, the graph-based HUGNets~\cite{hug} employs a hemispherical update mechanism compatible with continuous event transmission, significantly reducing prediction latency. Recently, Gehrig et al.~\cite{dense} proposed using parameterized B\'ezier curves to represent pixel trajectories in continuous time. Although this method often relies on deep networks to predict control points, its output consists of continuous functions rather than discrete frames, preserving the ability to query optical flow at an arbitrary timestamp.Despite the high accuracy of synchronous deep learning methods, their forced discretization of continuous event streams inherently contradicts the high temporal resolution of event cameras. In contrast, while asynchronous methods are more complex to design, they better align with the physical characteristics of event cameras, enabling low-latency, continuous-time motion estimation, a critical requirement for high-speed control applications.

\textbf{Event-based motion estimation}: Event cameras asynchronously detect per-pixel brightness changes, producing events that encode temporal information with high precision and low latency. Contrast Maximization (CMax) aligns events by warping them to a reference view and maximizing the variance of the resulting image, effectively recovering motion parameters through contrast optimization~\cite{Cmax1}. Complementary approaches, such as dispersion minimization~\cite{Dmin} and progressive event-to-map alignment scheme~\cite{STalig}, enhance robustness and efficiency by employing alternative loss functions and optimization strategies. Some techniques based on geometry establish direct relationships between events and motion parameters. For example, incidence relations between events and lines have been used for motion estimation~\cite{line1,line2}, though early methods often relied on external sensors (e.g., IMUs) for rotational velocity, limiting fully event-based solutions. Recent advances overcome this by developing solvers that jointly estimate rotational and translational velocities directly from event data. These approaches leverage event manifolds induced by line segments~\cite{line3} or normal flow~\cite{normflow} to recover full-DoF motion without external dependencies. Learning methods have emerged as a powerful alternative, utilizing neural networks to map event data to motion parameters~\cite{EMoflow}. Methods such as unsupervised learning frameworks~\cite{unsuper}, dense patch-based architectures~\cite{devo}, and event-frame fusion~\cite{RAMP} have been developed to enhance the applicability of learning-based approaches, albeit with hardware dependencies.

\section{Problem formulation}
\label{sec:Problem formulation}
In this section, we introduce the fundamentals of the differential epipolar constraints and formulate the basic framework for recovering the camera's angular and linear velocities from asynchronous optical flow.

\subsection{The decoupled differential epipolar constraints}
\label{subsec:The decoupled differential epipolar constraints}
\begin{figure}[htbp]
	\centering
	\includegraphics[width=0.35\linewidth]{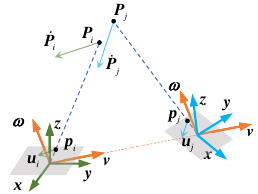}
	\caption{Geometric illustration of differential epipolar constraints. The camera's ego-motion (characterized by linear velocity $\boldsymbol{v}$ and angular velocity $\boldsymbol{\omega}$) induces relative 3D velocities ($\dot{\boldsymbol{P}}_i, \dot{\boldsymbol{P}}_j$) on static spatial points ($\boldsymbol{P}_i, \boldsymbol{P}_j$). The projection of these 3D velocities onto the image plane forms optical flow vectors ($\boldsymbol{u}_i, \boldsymbol{u}_j$) at the normalized image coordinates ($\boldsymbol{p}_i, \boldsymbol{p}_j$).}
	\label{fig:multdsfm}
\end{figure}

Consider a calibrated camera moving within a static scene, observing a set of 3D points $\{\boldsymbol{P}_i\}_{i=1}^N$ (as illustrated by $\boldsymbol{P}_i$ and $\boldsymbol{P}_j$ in Figure \ref{fig:multdsfm}). For the sake of clarity in the subsequent derivation, we temporarily omit the subscript and consider a generic 3D point $\boldsymbol{P} \sim (X, Y, Z)$ projected onto the image plane at point $\boldsymbol{p}$, represented in the camera coordinate system. We denote their relationship using normalized image coordinates as \( \boldsymbol{P} \sim Z \cdot \boldsymbol{p} \sim Z\cdot (p_x, p_y, 1) \). At a time instant, the camera undergoes motion with angular velocity \( \boldsymbol{\omega} \sim (w_x, w_y, w_z) \) and linear velocity \( \boldsymbol{v} \sim (v_x, v_y, v_z) \). The projection of this motion onto the image plane generates an optical flow vector \( \boldsymbol{u} \sim (u_x, u_y, 0) \) at the point \( \boldsymbol{p} \).

The velocity of \( \boldsymbol{P} \) in the camera coordinate system during the motion is:
\begin{equation}
	\dot{\boldsymbol{P}} = -[\boldsymbol{\omega}]_{\times} \boldsymbol{P} - \boldsymbol{v},
	\label{eq:mf3d}
\end{equation}
where $\dot{\boldsymbol{P}}$ denotes the differential of $\boldsymbol{P}$ with respect to time $t$, $[\boldsymbol{\omega}]_{\times}$ represents the skew-symmetric matrix of $\boldsymbol{\omega}$, explicitly written as
\[
[\boldsymbol{\omega}]_{\times} = \begin{bmatrix}
	0 & -w_z & w_y \\
	w_z & 0 & -w_x \\
	-w_y & w_x & 0
\end{bmatrix}.
\]

By differentiating both sides of $\boldsymbol{P} = \mathbf{Z} \cdot \mathbf{p}$, we obtain:
\begin{equation}
	\dot{\boldsymbol{P}} = \dot{Z}\cdot \mathbf{p} + Z \cdot \dot{\mathbf{p}}.
	\label{eq:mf2d}
\end{equation}

For the sake of brevity, we use the optical flow \( \boldsymbol{u} \) to replace the temporal derivative of the image points in subsequent sections. Combining Eq.\eqref{eq:mf3d} with Eq.\eqref{eq:mf2d} yields:
\begin{equation}
	-[\boldsymbol{\omega}]_{\times} Z\cdot \mathbf{p} - \mathbf{v} = \dot{Z} \cdot \mathbf{p} + Z \cdot \boldsymbol{u}.
	\label{eq:mf2&3d}
\end{equation}

By taking the dot-product with the vector $\boldsymbol{v} \times \boldsymbol{p} = [\boldsymbol{v}]_{\times} \boldsymbol{p}$ with both sides of Eq.\eqref{eq:mf2&3d}, we obtain the differential epipolar constraints under camera motion at a moment:

\begin{equation}
	\boldsymbol{u}^T [\boldsymbol{v}]_{\times} \boldsymbol{p} - \boldsymbol{p}^T [\boldsymbol{\omega}]_{\times} [\boldsymbol{v}]_{\times} \boldsymbol{p} = 0.
	\label{eq:dsfm}
\end{equation}

In the case of traditional frame cameras, which use only one frame of image point coordinates and optical flow, the solution typically employs an 8-point linear method and obtains the translational and angular velocities of the camera through singular value decomposition (SVD)~\cite{8pts}. We express the skew-symmetric matrix multiplications in Eq.\eqref{eq:dsfm} using their corresponding cross products, leading to the equivalent form:
\begin{equation}
	\boldsymbol{u} \cdot (\boldsymbol{v} \times \boldsymbol{p}) - \boldsymbol{p} \cdot (\boldsymbol{\omega} \times (\boldsymbol{v} \times \boldsymbol{p})) = 0.
	\label{eq:vecdsfm1}
\end{equation}

By employing the vector triple product and the cyclic property of the scalar triple product, Eq.\eqref{eq:vecdsfm1} can be simplified to the following form:
\begin{equation}
	(\boldsymbol{p} \times \boldsymbol{u} - (\boldsymbol{\omega} \cdot \boldsymbol{p}) \boldsymbol{p} + \boldsymbol{p} \cdot \boldsymbol{p} \cdot \boldsymbol{\omega}) \cdot \boldsymbol{v} = 0.
	\label{eq:vecdsfm2}
\end{equation}

 The linear velocity $\boldsymbol{v}$  is isolated as a linear variable through Eq.\eqref{eq:vecdsfm2}. This algebraic structure allows us to extend the constraint to asynchronous event streams, enabling the construction of efficient solvers.

\subsection{Asynchronous differential epipolar constraints}\label{subsec:Asynchronous differential epipolar constraints}

Exploiting the microsecond-level resolution of event cameras, it is common practice to model camera motion as constant over sufficiently short time intervals~\cite{line3}. We present the general form of differential epipolar constraints across different timestamps in this section, which is the foundation for the solver developed subsequently.

To incorporate multiple asynchronous events or observations (e.g., $\boldsymbol{p}_i$ and $\boldsymbol{p}_j$ shown in Figure \ref{fig:multdsfm}), we reintroduce the index $i$ to formulate a system of constraints based on their respective coordinates $\boldsymbol{p}_i$ and optical flow $\boldsymbol{u}_i$. Let $\boldsymbol{e}_i = \{\boldsymbol{p}_i, t_i\}$ denote the $i$-th event triggered by a static scene, comprising the normalized image coordinates $\boldsymbol{p}_i$ and the corresponding timestamp $t_i$. Suppose the optical flow $\boldsymbol{u}_i$ is extracted at these event coordinates. Furthermore, let $\boldsymbol{\omega}_i$ and $\boldsymbol{v}_i$ denote the instantaneous angular and linear velocities, respectively, defined in the local camera coordinate system at time $t_i$.

It is critical to note that all parameters in Eq.\eqref{eq:vecdsfm2} are represented in their respective camera coordinate systems. In this paper, the world coordinate system is aligned with the first camera coordinate system at the start of camera motion. Let $t_1=0$, and the timestamps of other events have a fixed deviation from it. 

Based on the assumption that the camera moves with constant angular and linear velocities in the world coordinate system within a brief temporal window, the corresponding values are denoted as $\boldsymbol{\omega}$ and $\boldsymbol{v}$, which satisfy the following relation with the instantaneous local velocities $\boldsymbol{\omega}_i$ and $\boldsymbol{v}_i$:

\begin{equation}
	\left\{
	\begin{aligned}
		\boldsymbol{\omega}_i &= \boldsymbol{\omega}, \\
		\boldsymbol{v}_i &= \boldsymbol{R}_i \boldsymbol{v},
	\end{aligned}
	\right.
	\label{eq:tmotion}
\end{equation}
where $\boldsymbol{R}_i$ denotes the rotation matrix between the world coordinate frame and the camera coordinate frame at time $t_i$, which is determined by $\boldsymbol{\omega}$ and time $t_i$. The differential epipolar constraints at multi-time can be expressed as:

\begin{equation}
	(\boldsymbol{p}_i \times \boldsymbol{u}_i - (\boldsymbol{\omega} \cdot \boldsymbol{p}_i) \boldsymbol{p}_i + \boldsymbol{p}_i \cdot \boldsymbol{p}_i \cdot \boldsymbol{\omega}) \boldsymbol{R}_i \boldsymbol{v} = 0.
	\label{eq:multsfm}
\end{equation}

 The geometric relationship between asynchronous event observations and the underlying continuous camera motion is formulated by Eq.\eqref{eq:multsfm}. The subsequent section will detail how we parameterize this system and design an efficient algebraic solver to accurately recover the global motion parameters $\boldsymbol{\omega}$ and $\boldsymbol{v}$.

\section{Egomotion Estimation}
\label{sec: Egomotion Estimation}
Building upon the theoretical foundations laid in the previous section, we develop 3 solvers to meet different practical requirements. Sec.\ref{subsec:miniev} introduces an exact formulation via singular value minimization. To ensure algebraic solvability, Sec.\ref{subsec:poly} employs a first-order rotation approximation using a Gr\"obner basis. Finally, Sec.\ref{subsec:Trunc} truncates higher-order terms of angular velocity, sacrificing large-rotation robustness for maximal computational speed.

\subsection{Eigenvalues Minimization Solver}
\label{subsec:miniev}
This subsection details an approach to solving Eq.\eqref{eq:multsfm}, where a singular value minimization algorithm decouples the problem into sequential optimizations for $\boldsymbol{\omega}$ and $\boldsymbol{v}$. Theoretically, this process requires only 5 events to achieve a solution. Under the multi-time epipolar differential epipolar constraints, Eq.\eqref{eq:multsfm} can be written as:
\begin{equation}
	\boldsymbol{A}(\boldsymbol{\omega}) \boldsymbol{v} = 0,
	\label{eq:simpcons}
\end{equation}
where $\boldsymbol{A}(\boldsymbol{\omega}) = (\boldsymbol{p}_i \times \boldsymbol{u}_i - (\boldsymbol{\omega} \cdot \boldsymbol{p}_i) \boldsymbol{p}_i + \boldsymbol{p}_i \cdot \boldsymbol{p}_i \cdot \boldsymbol{\omega}) \boldsymbol{R}_i $ is contingent upon the unknown $\boldsymbol{\omega}$. The rotation matrix $\boldsymbol{R}_i$, as defined by Rodrigues' rotation formula, can be explicitly expressed as follows:
\begin{equation}
	\boldsymbol{R}_i = \exp([t_i \boldsymbol{\omega}]_{\times}) = \mathbf{I} + \frac{\sin(\theta_i)}{\theta_i} [t_i \boldsymbol{\omega}]_{\times} + \frac{1 - \cos(\theta_i)}{\theta_i^2} [t_i \boldsymbol{\omega}]_{\times}^2,
	\label{eq:Rodrigues}
\end{equation}
where $\theta_i$ represents the rotation angle, defined as the magnitude of the vector $t_i \boldsymbol{\omega}$, and $t_i \boldsymbol{\omega} / \theta_i$ denotes the unit vector along the rotation axis.

 In Eq.\eqref{eq:simpcons}, owing to the inherent scale ambiguity in monocular vision, the magnitude of the linear velocity vector $\boldsymbol{v}$ is indeterminable, only its direction can be recovered, thus it possesses 2-DoF. Together with the 3-DoF from $\boldsymbol{\omega}$, there are a total of 5 unknown motion parameters to be estimated. Constraining these 5 unknowns requires at least 5 independent linear equations. By analyzing the sub-determinants of matrix $\boldsymbol{A}(\boldsymbol{\omega})$ and setting them to zero, it can be rigorously proven that a minimum of 5 point pairs are necessary to ensure a unique solution.

Thus the solution to the equation $\boldsymbol{A}(\boldsymbol{\omega}) \boldsymbol{v} = 0$ exists when the linear velocity is non-zero and the coefficient matrix $\boldsymbol{A}(\boldsymbol{\omega})$ possesses a non-trivial null space. To estimate the angular velocity $\boldsymbol{\omega}$, the problem is formulated as an optimization that minimizes the singular values of $\boldsymbol{A}(\boldsymbol{\omega})$, particularly targeting the square root of the smallest eigenvalue of $\boldsymbol{M}(\boldsymbol{\omega})$, which is defined as:
\begin{equation}
	\boldsymbol{M}(\boldsymbol{\omega}) = \boldsymbol{A}^T(\boldsymbol{\omega}) \boldsymbol{A}(\boldsymbol{\omega}).
	\label{eq:Matrix}
\end{equation}

Given that $\boldsymbol{M}$ is a real symmetric and positive semidefinite $3 \times 3$ matrix, the objective function is equivalent to minimizing its smallest eigenvalue $\lambda_{\min}$:
\begin{equation}
	\boldsymbol{\omega}^* = \arg \min_{\boldsymbol{\omega}} \lambda_{\min} (\boldsymbol{M}(\boldsymbol{\omega})).
	\label{eq:optimiz}
\end{equation}

The linear velocity $\boldsymbol{v}$ is subsequently obtained as the solution to the homogeneous polynomial system Eq.\eqref{eq:simpcons} .

\subsection{The Polynomial 5 Point Solver}
\label{subsec:poly}
Under asynchronous timing conditions, the use of an exact rotation matrix introduces complex, nonlinear constraints. To circumvent these difficulties for egomotion estimation, a Gr\"obner basis approach is employed here. A key step in this method is to replace the exact rotation matrix in Eq.\eqref{eq:Rodrigues} with its first-order approximation. This approximation reduces the problem to solving a system of polynomial equations, thereby enabling the application of computational algebraic geometry techniques. 
Under small rotation angles, the first-order approximation formula for the rotation matrix is:
\begin{equation}
	\mathbf{R}_i \doteq \mathbf{I} + [t_i \boldsymbol{\omega}]_{\times} = \begin{bmatrix}
		1 & -t_i w_z & t_i w_y \\
		t_i w_z & 1 & -t_i w_x \\
		-t_i w_y & t_i w_x & 1
	\end{bmatrix},
	\label{eq:appR}
\end{equation}
where $\mathbf{I}_{3 \times 3}$ is an identity matrix. The constraints are reformulated into a system of polynomial equations by substituting Eq.\eqref{eq:appR} into Eq.\eqref{eq:multsfm}. Due to the inherent scale ambiguity, only five unknowns need to be estimated. Following common practice, we therefore adopt the parameterization $\boldsymbol{v} \sim (v_x, v_y, 1)$ for the unscaled translation velocity estimation.This transformation yields a polynomial system comprising 27 monomials, with a total maximum degree of 3 and a maximum degree of 2 in the variable $\boldsymbol{\omega}$.

To construct the minimal solver, we arrange the coefficients of this polynomial system for 5 input event-correspondences into a matrix. The null space of this coefficient matrix provides the solution to the monomial vector $\boldsymbol{x}$. By applying Gaussian elimination to eliminate trivial rows, we obtain a condensed matrix $\mathbf{N}$.

Thus, the problem is formulated as finding the null vector of a $5 \times 27$matrix:

\begin{equation}
	\mathbf{N}_{5 \times 27} \boldsymbol{x} = 0,
	\label{RFOApoly}
\end{equation}
where 
\begin{equation}
	\begin{aligned}
		\boldsymbol{x} = & \left[ v_y w_x^2, \, w_x^2, \, v_x w_x w_y, \, v_y w_x w_y, \, w_x w_y, v_x w_x w_z, \, v_y w_x w_z, \, w_x w_z, \, v_y w_y^2,\right. \\
		& \left. w_y^2, \, v_x w_y w_z, \, w_z, \, v_y w_y w_z, \, w_z, \, v_x^2, \, v_y w_z^2, \, v_y w_z, \, v_x w_z^2, \, w_z^2, \, v_x v_y, \, 1 \right]^T.
	\end{aligned}
	\label{eq:27poly}
\end{equation}

The final solver, generated via ~\cite{larsson2017}, operates on a template of size $433 \times 473$, which yields 40 solutions. Notably, by setting $\boldsymbol{v} \sim (v_x, v_y, 1)$ , the velocity direction is constrained to a plane $z$ = 1 from the origin, which cannot represent pure lateral or vertical motions perpendicular to the $z$-axis (where $v_z$ = 0). However, the condition $v_z$ = 1 acts as a linear constraint that reduces the number of unknowns by one when constructing the elimination matrix, without increasing the degree of the equations. As a result, the final Gr\"obner basis becomes smaller, leading to improved numerical stability and faster solving speed. Furthermore, considering the decoupled angle and linear velocity equations in Eq.\eqref{eq:multsfm}, both $(v_x, v_y, 1)$ and its opposite $(-v_x, -v_y, -1)$ satisfy the system. Thus, reversing the signs of $v_x$ and $v_y$ in the process allows the representation of backward motion.

\subsection{Fast Solver via High-Order Truncation}
\label{subsec:Trunc}
Under the small-rotation assumption, the components of the rotation vector are considered small. Given this, the higher-order terms of $\boldsymbol{\omega}$ in Eq.\eqref{eq:27poly} become negligible compared to the lower-order ones. Truncating the cubic terms reduces the system to a set of 10 quadratic equations in 12 monomials, expressed as

\begin{equation}
	\mathbf{Q}_{5 \times 12} \boldsymbol{y} = 0,
	\label{eq:wTunc}
\end{equation}
where
\begin{equation}
	\begin{aligned}
		\boldsymbol{y} = [v_x w_x, v_y w_x, w_x, v_x w_y, v_y w_y, w_y, v_x w_z, v_y w_z, w_z, v_x, v_y, 1]^T. 
	\end{aligned}
	\label{eq:12poly}
\end{equation}

The monomial vector $\boldsymbol{y}$ (Eq.\eqref{eq:12poly}) corresponds to the lower-order part of $\boldsymbol{\omega}$ in the original vector $\boldsymbol{x}$ (Eq.\eqref{eq:27poly}). The matrix $\mathbf{Q}$ is the coefficient matrix of this truncated system. Eq.\eqref{eq:wTunc} is a polynomial system and we solve it using the Gr\"obner basis method. The generated automatic solver template size is $72 \times 82$. This approach is much faster than the method in Sec.\ref{subsec:poly} and reduces the number of solutions to 10.

\section{Experiment}
\label{sec:Experiment}

Following a consistent experimental setup, we evaluate the three solvers mentioned above: the eigenvalues minimization (\textbf{MiniEV}), the first-order approximation for rotation matrix (\textbf{ApproxRM}), the fast solver via high-order truncation of the angular velocity \textbf{$\boldsymbol{w}$} (\textbf{TruncAV}). The classical linear solver\cite{8pts} and the asynchronous matching-based method\cite{asynchronous1} serve as baseline algorithms and are compared in the subsequent sections of this chapter.

We adopt the same metric in Zhao et al.\cite{line3} to quantitatively assess the performance of the recovered angle velocities $\boldsymbol{\omega}_{\text{est}}$. Let $\boldsymbol{\omega}_{\text{gt}}$ denote the ground-truth angular velocity. The evaluation metric is defined as:
\begin{equation}
	\varepsilon_{\text{ang}} \left( \boldsymbol{\omega}_{\text{est}}, \boldsymbol{\omega}_{\text{gt}} \right) = \frac{\left\| \boldsymbol{\omega}_{\text{est}} - \boldsymbol{\omega}_{\text{gt}} \right\|}{\left( \left\| \boldsymbol{\omega}_{\text{est}} \right\| + \left\| \boldsymbol{\omega}_{\text{gt}} \right\| \right)}.
	\label{eq:angerror}
\end{equation}

The resulting error is confined to the interval $[0, 1]$, with values closer to zero indicating higher estimation accuracy. Similarly, for evaluating the recovered linear velocities, we use the angle deviation between the estimated and ground-truth vectors as the performance measure:
\begin{equation}
	\varepsilon_{\text{lin}} \left( \boldsymbol{v}_{\text{est}}, \boldsymbol{v}_{\text{gt}} \right) = \arccos \left( \frac{\boldsymbol{v}_{\text{est}}^T \boldsymbol{v}_{\text{gt}}}{\left\| \boldsymbol{v}_{\text{est}} \right\| \left\| \boldsymbol{v}_{\text{gt}} \right\|} \right),
	\label{eq:verror}
\end{equation}
where $\boldsymbol{v}_{\text{gt}}$ denotes the ground-truth linear velocity and $\boldsymbol{v}_{\text{est}}$ denotes the estimated linear velocity. This error metric provides a scale-invariant assessment of directional accuracy.

All experiments were conducted on a laptop equipped with an Intel(R) Core(TM) Ultra 7 255HX CPU and 32 GB of RAM. The system operated without dedicated GPU acceleration to maintain a consistent baseline for evaluating computational efficiency and convergence behavior across methods.

\subsection{Simulation}
\label{subsec:Simulation}
In this section, we conducted simulation experiments to evaluate the performance of the proposed solvers. The ground-truth angular velocity $\boldsymbol{\omega}$ and linear velocity $\boldsymbol{v}$ were sampled from uniform distributions spanning [$-$1/8, 1/8] rad/s and [$-$5, 5] m/s respectively. The event set was generated with a time interval of 0.5 seconds. To ensure sampling generality, we randomly generated three-dimensional points within a conical field of view characterized by a $45^{\circ}$angle and a depth range of [1, 20] meters. The corresponding optical flow was then derived based on the sampled true linear and angle velocities. All simulations utilized a virtual camera with a focal length $f$ = 400 pixels.

\textbf{Numerical Stability and Runtime.} The performance of the solvers is evaluated in terms of efficiency (runtime) and numerical stability on synthetic data free from observational noise. After 10,000 trials per solver, the median angular and linear velocity estimation errors are recorded. Given the propensity of the \textbf{MiniEV} to converge to local optima, it was initialized with random perturbations near the ground truth, its stability is further quantified by the success rate (\textbf{SR}) at two thresholds: \textbf{SR1} (error $<$ 0.01) and \textbf{SR2} (error $<$ 0.05). The median solution error is derived from the objective function value of the estimated solutions. The performance of the \textbf{MiniEV} is shown in Table\ref{tab:miniev_stability}, with a median solver runtime of 2.5 ms.

\begin{table}[htb]
	\centering
	\caption{Runtime, numerical stability, and success rates of \textbf{MINIEV} on noise-free synthetic data.}
	\label{tab:miniev_stability}
	\resizebox{\textwidth}{!}{%
	\begin{tabular}{@{}l c c c c c c@{}}
		\toprule
		\textbf{Formulation} & \textbf{Median $\mathbf{\epsilon_{\text{ang}}}$} & \textbf{Median $\mathbf{\epsilon_{\text{lin}}}(\mathbf{^\circ})$} & \textbf{SR1}(\%) & \textbf{SR2}(\%) & \textbf{Median objective} & \textbf{Runtime($\mathbf{\mu s}$)} \\
		\midrule
		\textbf{MiniEV} & $8.36 \times 10^{-3}$ & $2.45 \times 10^{-1}$ & 48.87 & 87.17 & $-4.19 \times 10^{-9}$ & $2.50 \times 10^3$ \\
		\bottomrule
	\end{tabular}%
	}
\end{table}

\begin{table}[htb]
	\centering
	\caption{Runtime, numerical stability of \textbf{ApproxRM} and \textbf{TruncAV} on noise-free synthetic data.}
	\label{tab:rfoa_wtrun_stability}
	\begin{tabular}{@{}l c c c c c@{}}
		\toprule
		\textbf{Formulation} & \textbf{Vars} & \textbf{Sols} & \textbf{Median $\mathbf{\epsilon_{\text{ang}}}$} & \textbf{Median $\mathbf{\epsilon_{\text{lin}}}(\mathbf{^\circ})$} & \textbf{Median runtime($\mathbf{\mu s}$)} \\
		\midrule
		\textbf{ApproxRM} & 5 & 40 & $4.28 \times 10^{-7}$ & $3.08 \times 10^{-6}$ & $2.77 \times 10^3$ \\
		\textbf{TruncAV} & 5 & 10 & $1.10 \times 10^{-3}$ & $2.66 \times 10^{-2}$ & $2.17 \times 10^2$ \\
		\bottomrule
	\end{tabular}
\end{table}

We evaluated the Gr\"obner basis solvers on standard data with image coordinates and optical flow generated under the first-order approximation rotation matrix. As shown in Table\ref{tab:rfoa_wtrun_stability}, the \textbf{ApproxRM} solver produced 40 polynomial solutions (\textbf{Sols}) with high precision and a runtime comparable to that of \textbf{MiniEV}. In contrast, \textbf{TruncAV}, which discards higher-order terms, exhibited significantly lower precision. Nevertheless, under the minimal configuration, its computational precision was on par with the minimal singular value method, while its computation time was 12.7 times faster than \textbf{ApproxRM} and over 11.5 times than \textbf{MiniEV} respectively.

\begin{figure}[htbp]
	\centering
	\includegraphics[width=0.9\linewidth]{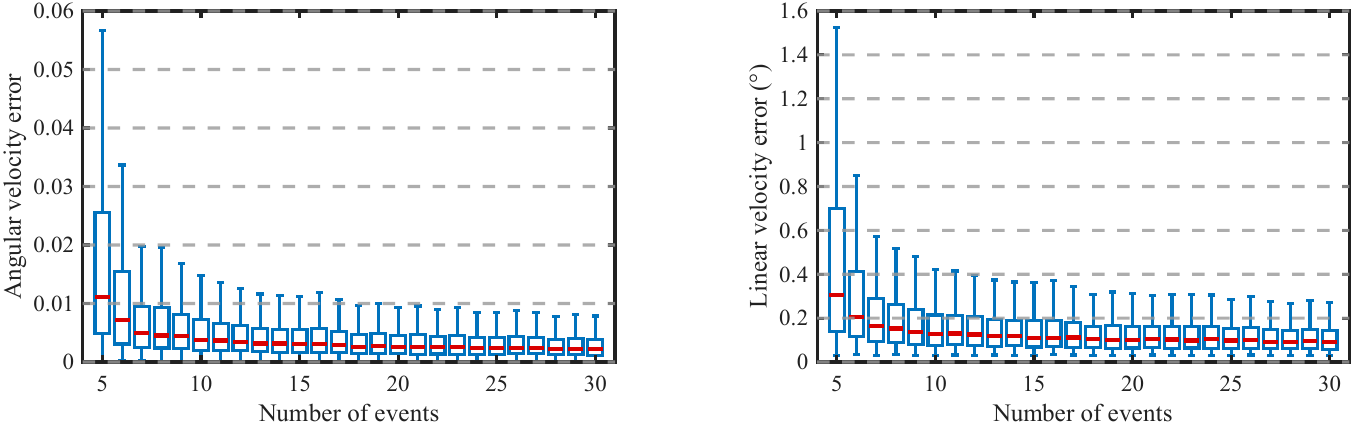}
	\caption{Statistical evaluation of estimation errors for angular velocity (left) and linear velocity (right) with respect to the number of events.}
	\label{fig:num_es}
\end{figure}

\textbf{Number of events.} To validate the robustness of the \textbf{MiniEV} under conditions of extreme data sparsity, we designed a quantitative evaluation experiment concerning the number of sampled events. In this experiment, the number of events used for estimation was incrementally increased from $5$ to $30$. For each sampling density, $1,000$ trials were conducted to obtain the error distribution data. The statistical distribution of errors is shown in Figure \ref{fig:num_es}. The red horizontal lines indicate the median values, while the upper and lower boundaries of the boxes represent the upper and lower quartiles, respectively. This visualization intuitively reflects the accuracy and stability of the algorithm. The left and right subplots correspond to the error distributions of angular velocity and linear velocity.

Experimental results demonstrate that the \textbf{MiniEV} algorithm exhibits significant convergence characteristics. When the number of input events is low, the estimation error decays exponentially with the increase in the number of events. Under the extremely sparse condition of $N=5$, although the dispersion of the error distribution is relatively large, the median error remains within an acceptable range, verifying the fundamental feasibility of the algorithm. As $N$ increases to $15$, the median errors for both angular and linear velocities decrease by approximately $77\%$, accompanied by a significant contraction in variance. When $N>20$, the marginal gain in accuracy begins to diminish, and the variations in median errors for angular and linear velocities across different event counts are bounded by $4.0\times10^{-4}$ and $1.4\times10^{-3}$, respectively. These minimal fluctuations confirm that the error distributions have stabilized, indicating that the algorithm has reached a steady state.

\textbf{Noise Resilience Analysis.} To increase realism, Gaussian noise was introduced to the simulated data. Specifically, a standard deviation of $\sigma = 5 \sim 25$ pixel was applied to the event coordinates, $\sigma = |\boldsymbol{u}|/40 \sim |\boldsymbol{u}|/8$ pixel/s to the optical flow coordinates, and $\sigma = 0.04 \sim 0.2 s$ (2$t_{\text{max}}/25 \sim 2 t_{\text{max}}/5 $) to the timestamps.
We compared the proposed asynchronous method with the traditional 8-point algorithm (\textbf{Linear-8p}) \cite{8pts}. The 8-point algorithm disregards timestamp information and aggregates events and optical flow within a $0.5 s$ window into a single frame. The \textbf{MiniEV} solver was implemented in both 5-point (\textbf{MiniEV-5p}) and 8-point (\textbf{MiniEV-8p}) configurations, while the Gr\"obner basis-based solvers utilized 5 points. Each solver was executed 1,000 times, with the standard deviation of the noise scaled from different levels.

\begin{figure}[h]
	\centering
	\captionsetup[subfigure]{labelformat=empty}
	
	\subfloat[\small (a) Robustness to Pixel Noise ]{
		\includegraphics[width=0.95\linewidth]{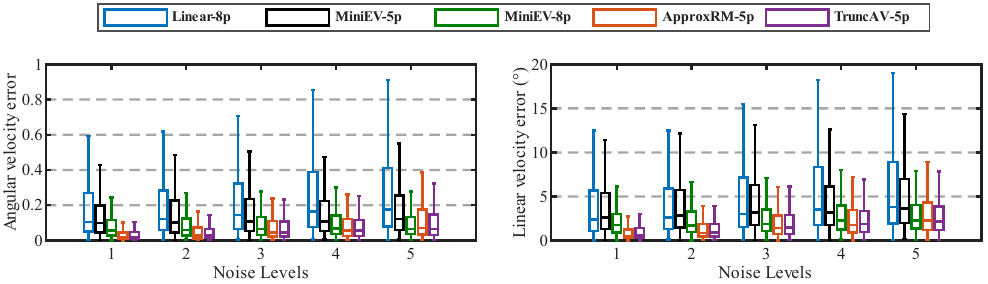}
		\label{fig:img1}
	}
	\vspace{-0.3cm} \\
	
	\subfloat[\small (b) Robustness to Optical Flow Noise]{
		\includegraphics[width=0.95\linewidth]{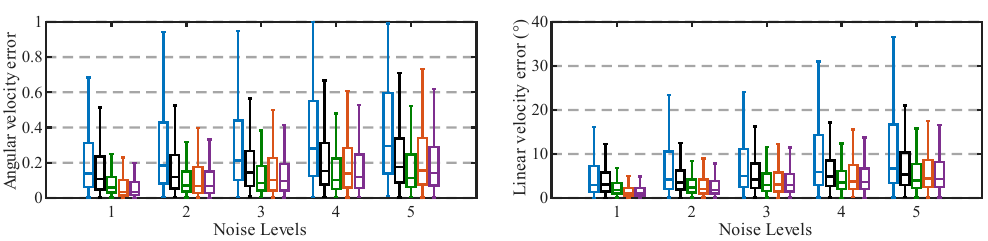}
		\vspace{-0.5cm}
		\label{fig:img4}
	}
	\vspace{-0.3cm} \\
	
	\subfloat[\small (c) Robustness to Timestamps]{
		\includegraphics[width=0.95\linewidth]{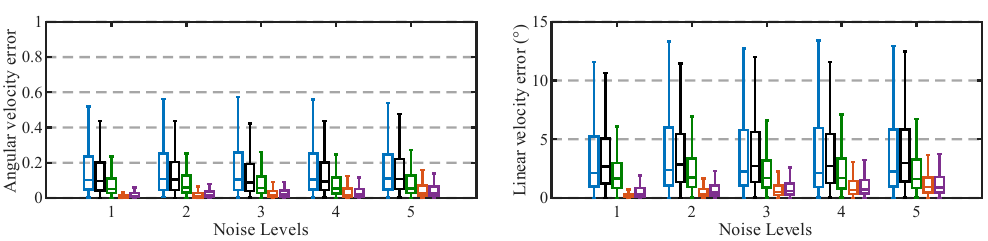}
		\label{fig:img2}
	}
	
	\caption{The results on synthetic data demonstrate the relationship between errors of various factors. The layout compares angular error and linear velocity error across three different noise factors. (a) Robustness to Pixel Noise.($\sigma = 5\sim25$ pixel) (b) Robustness to Optical Flow Noise.($\sigma = |\boldsymbol{u}|/40\sim|\boldsymbol{u}|/8$) (c) Robustness to Timestamps.( $\sigma = 0.04\sim0.2 s$ (2$t_{\text{max}}/25\sim2t_{\text{max}}/5 $))}
	\label{fig:six_images}
\end{figure}

The resulting solution errors are summarized in Figure \ref{fig:six_images}. The \textbf{Linear-8p} consistently exhibits lower accuracy than the asynchronous approach, and is even outperformed by the \textbf{MiniEV-5p} given reasonable initial values. This result underscores the importance of explicit temporal modeling for improving the accuracy of egomotion estimation in event cameras. \textbf{MiniEV-8p} achieved the highest accuracy, which we attribute to its utilization of a larger number of events. Interestingly, \textbf{TruncAV} demonstrated marginally more robust performance than the \textbf{ApproxRM} algorithm under high-noisy conditions. The performance trade-off between \textbf{ApproxRM} and \textbf{TruncAV} is clearly demonstrated in the pixel error and optical flow extraction error experiments, as illustrated in Figure \ref{fig:six_images}. Under low-noise conditions, \textbf{ApproxRM} exhibits a significantly lower median error and superior error convergence compared to \textbf{TruncAV}. This advantage likely stems from \textbf{ApproxRM}'s retention of higher-order terms, which more accurately capture the differential epipolar geometric constraints under well-conditioned scenarios, thereby maintaining higher estimation fidelity and stability. However, as the noise level increases, the performance of \textbf{ApproxRM} degrades rapidly and is eventually surpassed by \textbf{TruncAV}. This reversal suggests that the higher-order terms in \textbf{ApproxRM} may render it more sensitive to data corruption and noise. In contrast, by truncating the higher-order terms associated with infinitesimal rotational velocities, \textbf{TruncAV} demonstrates stronger robustness under high-error conditions, resulting in better adaptability in challenging and highly uncertain scenarios, a behavior that is also corroborated in real experiments.

Experimental results on temporally noisy synthetic data indicate that performance of algorithms is influenced differently by temporal noise compared to other noise types. \textbf{Linear-8p} lacks an explicit temporal model and exhibits nearly constant solution errors. In contrast, the Gr\"obner basis-based solver shows low sensitivity to temporal perturbations and achieves superior accuracy. When utilizing the same number of events, \textbf{MiniEV-8p} maintains a significant accuracy advantage over \textbf{Linear-8p}. A notable trade-off is observed with \textbf{MiniEV-5p}: it achieves higher accuracy in angular velocity estimation compared to \textbf{Linear-8p}, but this is offset by a slight reduction in linear velocity estimation accuracy. This trade-off is attributed to the sequential estimation process of \textbf{MiniEV-5p}. As shown in Eq.\eqref{eq:tmotion}, the angular velocity is invariant to the coordinate system and thus relatively immune to temporal errors. However, the linear velocity components are intrinsically linked to the timeline and the angular velocity magnitude, making them highly correlated with temporal errors. Therefore, even minor inaccuracies in the angular velocity estimation inevitably propagate and are amplified during the subsequent linear velocity estimation. Conversely, the unified cost function used in \textbf{Linear-8p} facilitates concurrent optimization of all parameters, enabling implicit error compensation and resulting in a solution that is balanced overall, albeit not optimal for any single parameter.

\textbf{Comparison with point-matching method.} Furthermore, we evaluated the recent advancements in asynchronous 5-point matching solvers (\textbf{AsynTR})\cite{asynchronous1}. Predicated on acquiring matched data within an asynchronous spatiotemporal space, this approach utilizes 5 asynchronously matched point correspondences to estimate angular and linear velocities. Under the K1, A2 configuration and problems with m=2, n=5 described in \cite{asynchronous1}, the generated solver requires an elimination template size of $190 \times 210$. Concurrently, by truncating higher-order angular velocity terms, we derived a more compact solver (\textbf{TruncATR}) characterized by an elimination template of $72 \times 82$. 

\begin{figure}[t]
	\centering
	\captionsetup[subfigure]{labelformat=empty}
	
	\subfloat[\small (a) Robustness to Pixel Noise ]{
		\includegraphics[width=0.75\linewidth]{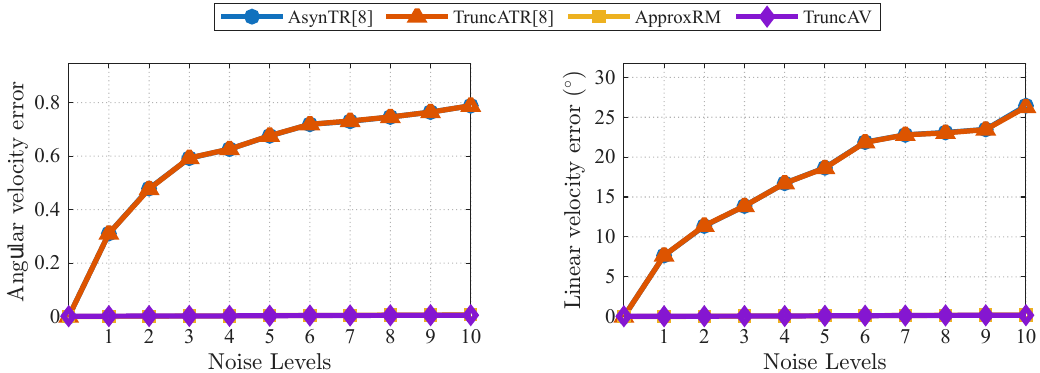}
		\label{fig:polypix}
	}
	\vspace{-0.3cm} \\
	
	\subfloat[\small (b) Robustness to Optical Flow and Matching Noise]{
		\includegraphics[width=0.75\linewidth]{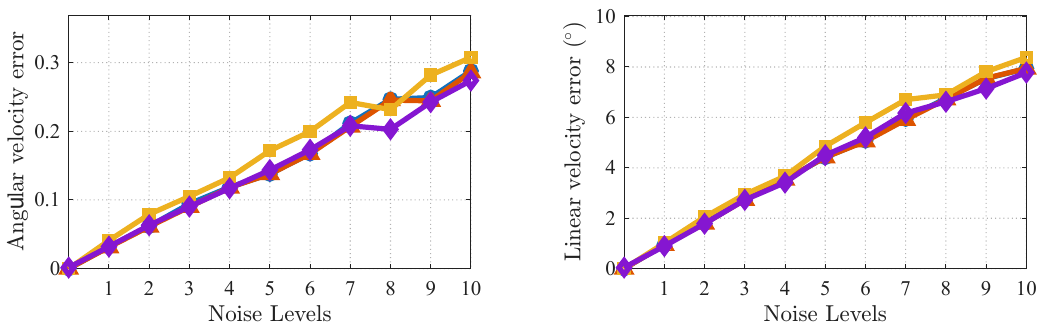}
		\vspace{-0.5cm}
		\label{fig:polyopfl}
	}
	\vspace{-0.3cm} \\
	
	\subfloat[\small (c) Robustness to Timestamps]{
		\includegraphics[width=0.75\linewidth]{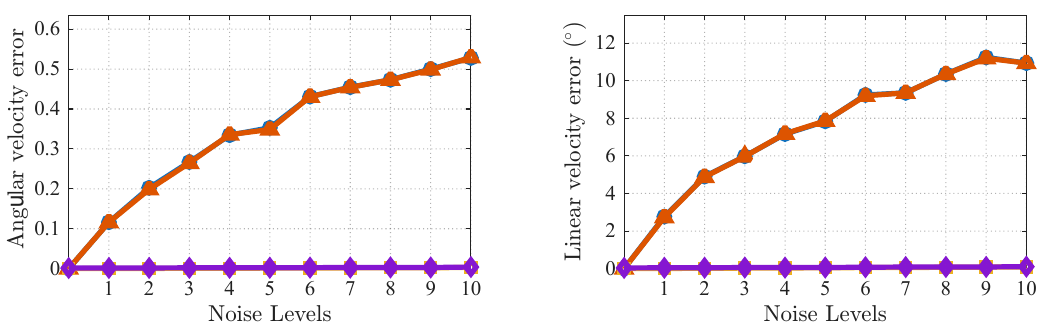}
		\label{fig:polyt}
	}
	
	\caption{Robustness evaluation of various asynchronous minimal solvers under multimodal noise on synthetic data. The layout compares the angular (left) and linear (right) velocity estimation errors of point-matching-based solvers (\textbf{AsynTR}, \textbf{TruncATR}) and optical-flow-based solvers (\textbf{ApproxRM}, \textbf{TruncAV}) across three distinct noise factors: (a) Pixel coordinate quantization noise.($\sigma = 0.2 \sim 2$ pixel) (b) Optical flow extraction and equivalent matching noise.($\sigma = |\boldsymbol{u}|/40 \sim |\boldsymbol{u}|/4$) (c) Temporal timestamp jitter.( $\sigma = 0.002 \sim 0.02 s$ (2$t_{\text{max}}/25 \sim 2 t_{\text{max}}/5 $))}
	\label{fig:poly_images}
\end{figure}

Following the aforementioned experimental setup, we set the focal length to $f = 400$ pixel. To strictly evaluate the accuracy disparity among polynomial minimal solvers, varying levels of Gaussian noise were injected into the pixel coordinates (with a baseline error of $\sigma = 0.2 $ pixel), optical flow, and timestamps. To enforce a mathematically fair comparison between optical flow and discrete coordinate matching, spatial noise was initially modeled on the first set of points. The noise scales were explicitly formulated to be proportional to either the discrete spatial distance $|\mathbf{p}_1 - \mathbf{p}_2|/f$ or the optical flow magnitude $|\mathbf{u}|/f$. Timestamp perturbations were bounded by $1/25$ of the maximum temporal interval across the five events. To maintain absolute equivalence via error propagation, the induced noise in the point-matching algorithms was evenly distributed across the two discrete observations. Each solver was executed 1,000 times, and the median solution errors are summarized in Figure \ref{fig:poly_images}.

As illustrated in Figure \ref{fig:poly_images}, \textbf{AsynTR} demonstrates acute sensitivity to both pixel and temporal noise. As the pixel noise level increases, its estimation error diverges rapidly, ultimately being outperformed by the optical flow-based asynchronous minimal solvers. Algebraically, this instability is rooted in the multiplication of two rotation matrices within its discrete epipolar constraint, which induces highly nonlinear error coupling and severely destabilizes the polynomial system under perturbations. However, minimal solvers relying on optical flow exhibit a significantly higher sensitivity to numerical inaccuracies in the optical flow estimates themselves---an effect that slightly surpasses the impact of raw event coordinate extraction errors on feature-point-based matching algorithms. Furthermore, we observe that solvers utilizing full polynomial approximations are noticeably more susceptible to noise corruption compared to those truncating higher-order angular velocity terms. By discarding these higher-order terms, the elimination template size and the number of feasible solutions are systematically reduced, rendering the solver both computationally more efficient and robust. This phenomenon is also corroborated in the \textbf{TruncATR} solver, which achieves accelerated resolution speeds with negligible degradation in accuracy. In conclusion, the observations indicate that optical flow-based egomotion recovery is intrinsically more resilient against temporal jitter and pixel quantization errors; nonetheless, this resilience fundamentally relies on the availability of highly precise optical flow measurements from the front-end.

\subsection{Real experiments}
\label{subsec:Real experiments}
In this section, we evaluate the performance of various solvers in a real-world scenario using the public M3ED dataset~\cite{M3ED}. It is worth noting that comprehensive real-world benchmarking for asynchronous egomotion is currently constrained by the extreme scarcity of datasets providing high-fidelity, asynchronous optical flow ground truth. Consequently, the real-world evaluation is exclusively conducted on the selected sequences of the M3ED dataset, where reliable flow references can be established. The experiments focus on the \mthreeedseq{} sequence, in which intervals that approximately satisfy the constant velocity and small rotation assumptions. We obtain the optical flow ground truth following the methodology of the $T^2CEF$ dataset~\cite{ttcef}. In contrast, the corresponding pose ground truth in this paper, characterized by a higher temporal resolution, was acquired via spline interpolation to facilitate rigorous algorithm benchmarking.

\begin{figure}[h]
	\centering
		\subfloat[Gray-scale image\label{fig:time}]{
		\includegraphics[width=0.32\linewidth]{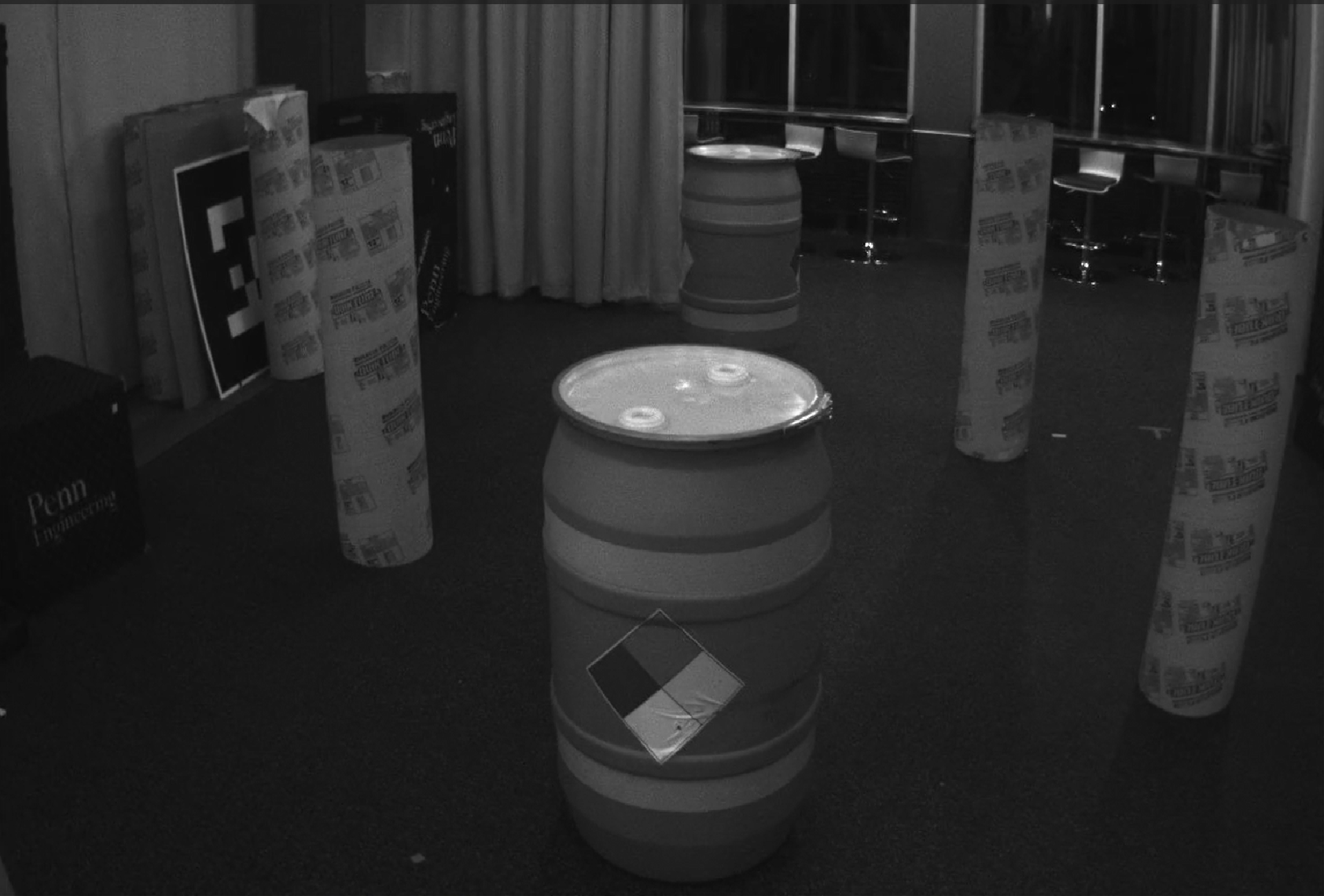}
	}
	\hfill
	\subfloat[2D event projection\label{fig:2d}]{
		\includegraphics[width=0.32\linewidth]{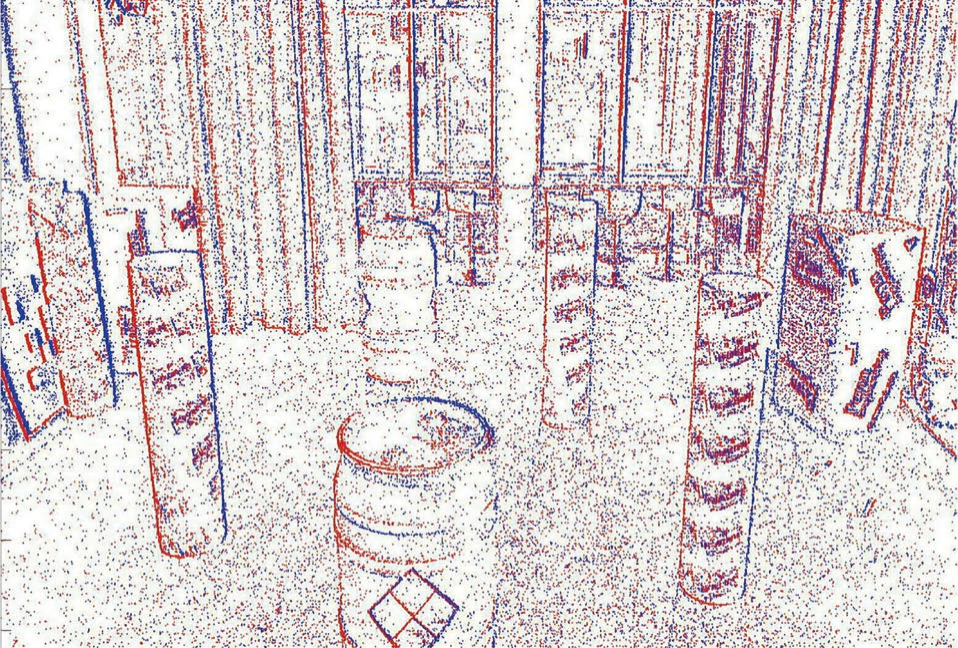}
	}
	\hfill
	\subfloat[Real events optical flow\label{fig:realflow}]{
		\includegraphics[width=0.30\linewidth]{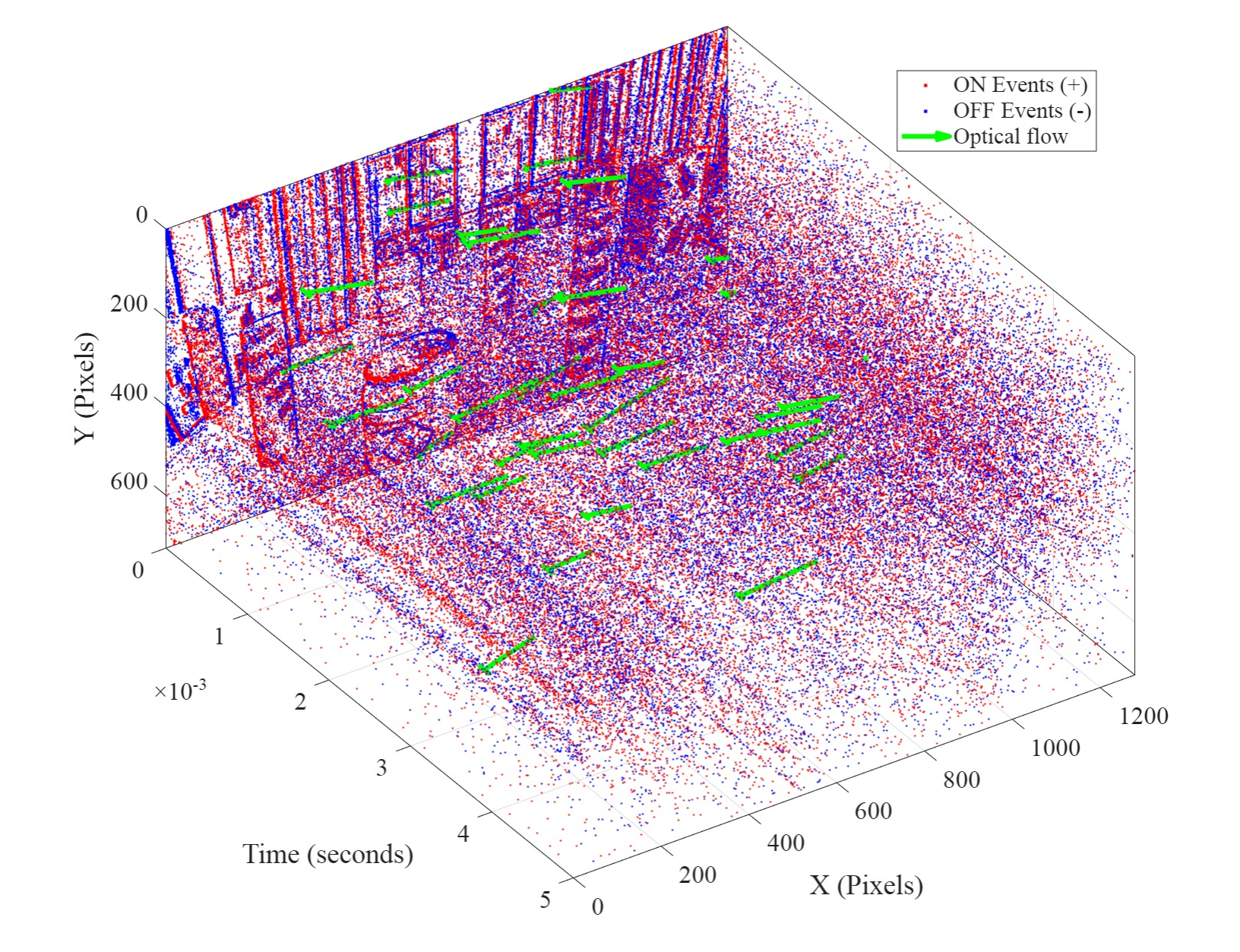}
	}
	
	\caption{Optical flow visualization for the \mthreeedseq{} sequences from the M3ED dataset. (a) A grayscale image from M3ED (only for visual reference). (b) Event streams accumulated over a 5ms interval and projected onto the image plane. (c) A three-dimensional visualization of the spatiotemporal event surface (polarity: red=positive, blue=negative), projected onto the XY plane, with a sparsely sampled set of optical flow vectors overlaid (green arrows).}
	\label{fig:events_flow}
\end{figure}

Camera motion is estimated every 5 ms. The asynchronous solver uses generated optical flow data calculated from every 1 ms event data, which is then calculated in the same time window for comparison against the traditional synchronous \textbf{Linear-8p} method. The estimation accuracy for linear and angular velocities is then quantitatively assessed using this ground truth with Eq.\eqref{eq:angerror} and \eqref{eq:verror}. As shown in Figure \ref{fig:events_flow}, gray-scale image from the real dataset is used to represent the visual scene. To illustrate the imaging characteristics of the event camera, event streams within a 5ms interval are projected onto a single image(Figure \ref{fig:events_flow}(b)). Additionally, the spatiotemporal surface of the event stream, together with the downsampled optical flow ground truth, is visualized in a three-dimensional coordinate system(Figure \ref{fig:events_flow}(c)).

\begin{table}[htbp]
\centering
\caption{Real-world experiment results. We report the median errors for $\varepsilon_{\text{ang}}$ and $\varepsilon_{\text{lin}}$, bold entries indicate the best performance (lowest error or shortest runtime) across all compared methods in each column, while underlined entries denote the second-best performance.}
\label{tab:real_world_results} 
\begin{tabular}{c c c c}
	\toprule
	Algorithm & \begin{tabular}{@{}c@{}} Median \\ $\varepsilon_{\text{ang}}$\end{tabular} & 
	\begin{tabular}{@{}c@{}} Median \\ $\varepsilon_{\text{lin}}$ (${}^\circ$)\end{tabular} & 
	\begin{tabular}{@{}c@{}} Runtime \\ (ms)\end{tabular} \\
	\midrule
	\textbf{Linear-8p}              & 0.5360  & 21.6988  & \textbf{0.094} \\
	\textbf{MiniEV-5p}              & 0.3312  & 10.3753  & 4.702 \\
	\textbf{MiniEV-8p}              & 0.4004  & 8.2937   & 20.806  \\
	\textbf{TruncAV}                & 0.1901  & 5.5909   & \underline{0.194} \\
	\textbf{TruncAV} + \textbf{MiniEV-8p}    & \textbf{0.1242}  & \textbf{3.5147}   & 18.203  \\
	\textbf{TruncAV} + \textbf{MiniEV-5p}                & \underline{0.1811}  & \underline{4.7654}   & 1.425 \\
	\bottomrule
\end{tabular}
\end{table}

The fact that \textbf{ApproxRM} fails to converge in real-world experiments stems from the instability in Gr\"obner basis determination, critically undermined by accumulated floating-point rounding errors during large-scale computation. Additionally, we attempted to benchmark the asynchronous 5-point matching solvers (\textbf{AsynTR} and \textbf{TruncATR}) in this real-world setup by generating discrete point correspondences via optical flow integration. However, due to the inevitable accumulation of real-world flow extraction noise and the error amplification during the multiplication of sequential rotation matrices (as analyzed in Sec.\ref{subsec:Simulation}), the coordinate perturbations far exceeded the solvers' algebraic tolerance. Consequently, \textbf{AsynTR} and \textbf{TruncATR} failed to converge to any physically meaningful solutions, precluding their inclusion in the quantitative comparison. The remaining results of real-world experiment are presented in Table\ref{tab:real_world_results}, \textbf{Linear-8p} achieves the fastest runtime due to its minimal number of SVD operations, but exhibits significant estimation errors in the solution. \textbf{TruncAV} demonstrated the best performance among the individual algorithms, achieving the highest accuracy within the second-shortest computation time, in which the solution accuracy was significantly superior to that of the other algorithms. The \textbf{MiniEV} algorithm often converges to local optima during iterative optimization, resulting in slightly higher errors compared to the \textbf{TruncAV} method. All algorithms presented in this paper achieve higher accuracy than \textbf{Linear-8p}, which is additionally influenced by asynchronous timing effects. Finally, we employed \textbf{TruncAV} to provide initial values and optimized the estimates using both the \textbf{MiniEV-8p} and \textbf{MiniEV-5p} methods. Both approaches achieved a substantial improvement in the accuracy of linear and angular velocity estimation, along with a significant reduction in the computational time required for iterative optimization. While  \textbf{MiniEV-8p} yielded a greater enhancement in temporal precision, the 5-point method demonstrated a shorter solution time.

\begin{figure}[htbp]
	\centering
	\includegraphics[width=0.9\linewidth]{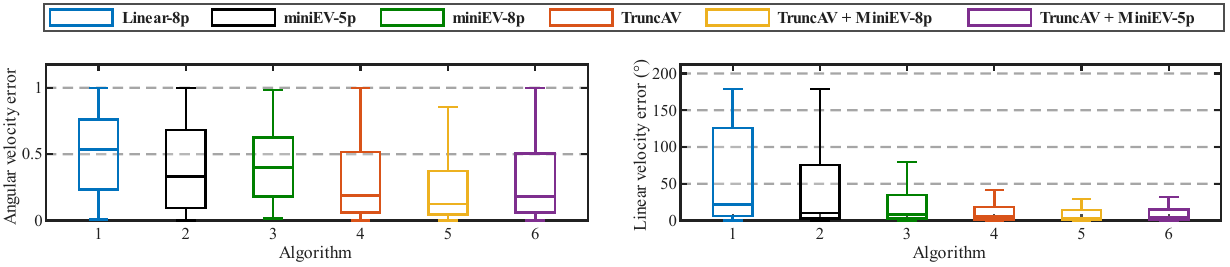}
	\caption{Statistical distribution of angular and linear velocity estimation errors for various algorithms evaluated on the real-world sequence.}
	\label{fig:realbox}
\end{figure}

To provide a more comprehensive evaluation of algorithmic robustness in complex real-world scenarios, we visualize the statistical distribution of estimation errors on the \mthreeedseq{} sequence using boxplots, as depicted in Figure \ref{fig:realbox}.

As illustrated in Figure \ref{fig:realbox}, the boxplots provide a clear visualization of the error dispersion for both angular and linear velocity estimation across the six evaluated algorithms. \textbf{Linear-8p} exhibits the highest median errors and the widest interquartile range, particularly in linear velocity estimation where the error distribution stretches extensively. This visually confirms the geometric degradation and temporal aliasing caused by forcing asynchronous event data into discrete frames. While iterative methods (\textbf{MiniEV-5p} and \textbf{MiniEV-8p}) manage to reduce the median errors compared to the baseline, their distributions remain highly dispersed with prominent whiskers. This indicates a continuing susceptibility to local optima and instability when exposed to unmodeled real-world noise. In contrast, the \textbf{TruncAV} method significantly compresses the error distribution. By algebraically truncating high-order noise-sensitive terms, it achieves a markedly narrower interquartile range, demonstrating enhanced numerical stability. Ultimately, the visual data strongly supports the superiority of the hybrid coarse-to-fine strategies. Both \textbf{TruncAV + MiniEV-8p} and \textbf{TruncAV + MiniEV-5p} deliver the most robust performance. Their respective boxplots are the most concentrated among all evaluated methods, effectively suppressing extreme outliers while achieving the lowest median errors. This rigorously substantiates the conclusion that initializing exact iterative solvers with robust algebraic approximations yields optimal stability and accuracy in dynamic, real-world environments.

\section{Conclusion}
\label{sec: Conclusion}

To address the fundamental challenge of estimating full-DoF camera motion from asynchronous event camera data streams, this paper decouples angular and linear velocities within the differential epipolar constraint equation and formally models the differential epipolar constraint for asynchronous observations. Building upon this formulation, we derive three minimal 5-point solvers specifically tailored for asynchronous event streams. Experimental results validate the benefits of explicitly modeling temporal asynchrony. Compared to traditional synchronous approaches (the 8-point algorithm) that disregard precise timestamp information, the proposed asynchronous framework demonstrates superior estimation accuracy and robustness against pixel, optical flow, and timestamp noise.We further investigate the trade-off between accuracy and computational efficiency: while the iterative optimization solver achieves theoretical optimality under ideal conditions, the polynomial-based Gr\"obner basis solver yields significant efficiency gains with negligible loss in accuracy. Notably, the adoption of a hybrid coarse-to-fine solving strategy in real-world experiments achieves an optimal balance between performance and efficiency. Despite these encouraging results, several avenues for future research merit further exploration. First, while the presented method performs exceptionally well in static environments, extending the differential epipolar framework to handle dynamic scenes containing moving objects remains a challenging open problem. Second, while the presented solvers demonstrate strong algebraic resilience, the final estimation accuracy remains heavily contingent upon the precision of the front-end optical flow extraction. Consequently, developing high-fidelity, purely asynchronous optical flow estimation algorithms represents a critical direction for future research.

\section*{Acknowledgements}
This work was supported by  the National Natural Science Foundation of China (Grant No. 12372189)  and the Science and Technology Innovation Program of Hunan Province (Grant No. 2025RC1045).

\bibliographystyle{unsrt}
\bibliography{References}

\end{document}